%% file: main.tex
\documentclass[10pt,twocolumn,letterpaper]{article}

\usepackage[final]{cvpr}      

\usepackage[misc]{ifsym} 
\usepackage[accsupp]{axessibility}  


\definecolor{cvprblue}{rgb}{0.21,0.49,0.74}
\usepackage[pagebackref,breaklinks,colorlinks,allcolors=cvprblue]{hyperref}

\def\paperID{13165} 
\def\confName{CVPR}
\def\confYear{2025}

\title{MTADiffusion: Mask Text Alignment Diffusion Model for Object Inpainting}

\author{
   Jun Huang\textsuperscript{\rm 1,\dag},
   Ting Liu\textsuperscript{\rm 1,\dag,\Letter},
   Yihang Wu\textsuperscript{\rm 2},
   Xiaochao Qu\textsuperscript{\rm 1},
   Luoqi Liu\textsuperscript{\rm 1},
   Xiaolin Hu\textsuperscript{\rm 3,\Letter}
\\ 
\small \textsuperscript{\rm 1} MT Lab, Meitu Inc, Beijing 100083, China \\
\small \textsuperscript{\rm 2} National University of Singapore \\
\small \textsuperscript{\rm 3} Department of Computer Science and Technology, \\
\small BNRist, IDG/McGovern Institute for Brain Research, \\
\small Tsinghua University, Beijing 100084, China \\
\small \texttt{\{hj13, lt, qxc, llq5\}@meitu.com} \\
\small \texttt{yihang\_wu@u.nus.edu} \\
\small \texttt{xlhu@tsinghua.edu.cn} \\
\vspace{1mm}
\small \textsuperscript{$\dagger$}Joint first authors. \small \hspace{3mm} \textsuperscript{\Letter\ }Joint corresponding authors.
\vspace{-0.75cm}
}

\begin{document}
\maketitle
\input{sec/0_abstract}    
\input{sec/1_intro}
\input{sec/2_related_work}
\input{sec/3_method}
\input{sec/4_exp}
\input{sec/5_conclusion}
{
    \small
    \bibliographystyle{ieeenat_fullname}
    \bibliography{main}
}


\end{document}

%% file: sec/0_abstract.tex
\begin{abstract}
Advancements in generative models have enabled image inpainting models to generate content within specific regions of an image based on provided prompts and masks. However, existing inpainting methods often suffer from problems such as semantic misalignment, structural distortion, and style inconsistency. In this work, we present MTADiffusion, a Mask-Text Alignment diffusion model designed for object inpainting. To enhance the semantic capabilities of the inpainting model, we introduce MTAPipeline, an automatic solution for annotating masks with detailed descriptions. Based on the MTAPipeline, we construct a new MTADataset comprising 5 million images and 25 million mask-text pairs. Furthermore, we propose a multi-task training strategy that integrates both inpainting and edge prediction tasks to improve structural stability. To promote style consistency, we present a novel inpainting style-consistency loss using a pre-trained VGG network and the Gram matrix. Comprehensive evaluations on BrushBench and EditBench demonstrate that MTADiffusion achieves state-of-the-art performance compared to other methods.
\end{abstract}

%% file: sec/1_intro.tex
\section{Introduction}
\label{sec:intro}

\begin{figure}
    \centering
    \includegraphics[width=0.45\textwidth]{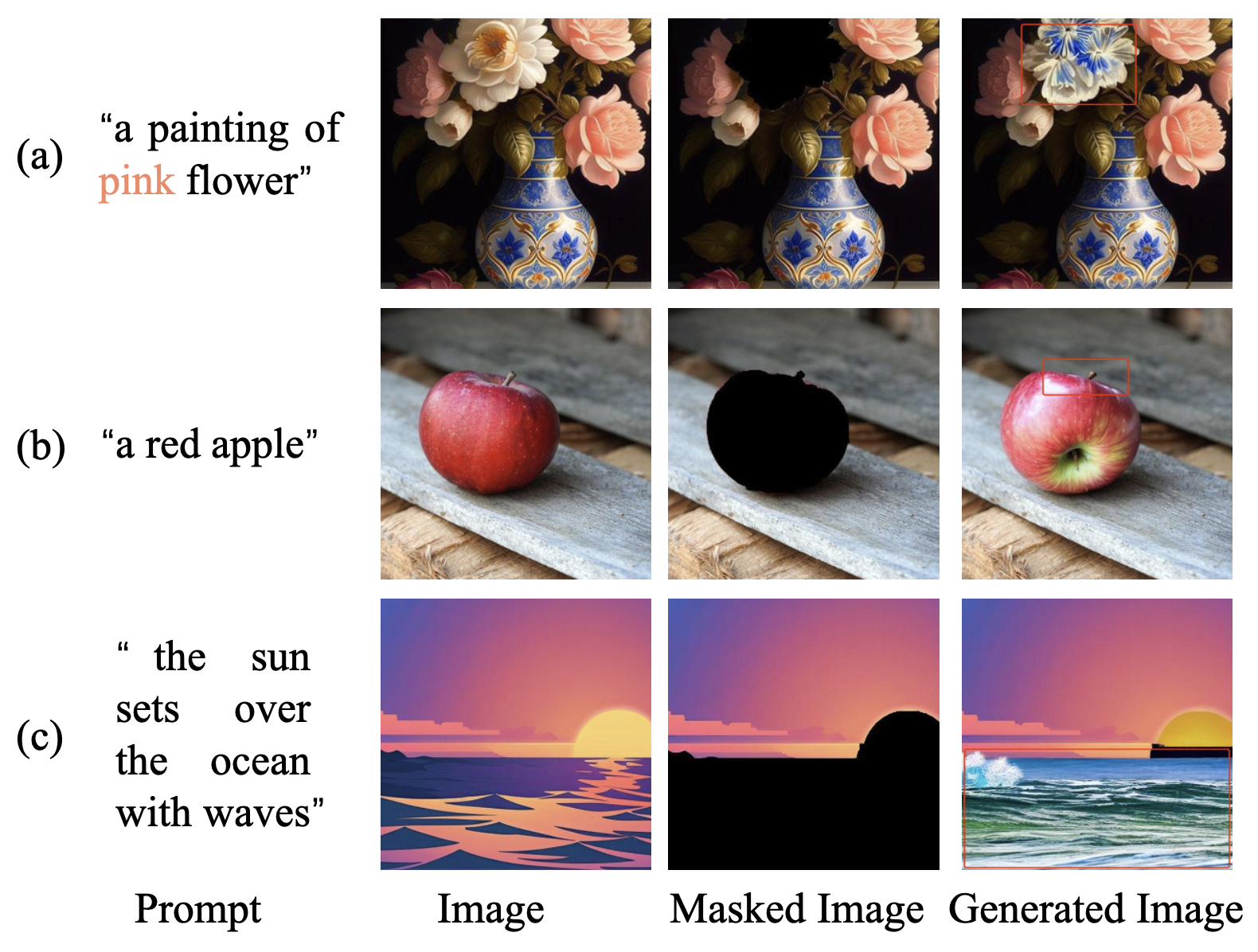}
        \caption{\textbf{Common issues in inpainting tasks.} (a) semantic misalignment. (b) structural distortion. (c) style inconsistency.}
    \label{fig:issues}
\end{figure}

With the progress of diffusion models, generative models are now capable of producing a wide variety of content. The image inpainting task aims to generate content within local regions of an original image based on the provided prompt and mask. Current challenges in this task as shown in Figure ~\ref{fig:issues} include the following: (a) \textbf{Semantic misalignment} \cite{xie2023smartbrush, zhuang2023task}: The generated content may not align semantically with the input prompt. For instance, the model might produce content that does not match the specifications of the prompt or tends to complete the masked area with background context rather than generating the intended new object. (b) \textbf{Structural distortion} \cite{rombach2022high}: The structural details of generated objects are distortion, such as disorganized limbs or fuzzy features. (c) \textbf{Style Inconsistency}: Even when the content of the generated object is reasonable, style inconsistency remains an issue. Specifically, aspects such as hue, texture, and lighting of the generated object are often inconsistent with those of the original image, resulting in unnatural visual outcomes.

Studies have focused on improving the semantic capabilities of inpainting models. SDI \cite{rombach2022high} trains the UNet with random mask and image caption pairs, where the image caption describes the entire image rather than the content of the masked region, causing semantic misalignment during the training stage. To address this, SmartBrush \cite{xie2023smartbrush} and PowerPaint \cite{zhuang2023task} align the object mask with text descriptions using the training split of OpenImages v6 \cite{OpenImages2}, which has segmentation masks and the corresponding labels. However, simple labels are far from sufficient to describe the mask region in detail, and the amount of labeled data cannot satisfy the demand for large model training. BrushNet \cite{ju2024brushnet} uses Grounded-SAM \cite{kirillov2023segany, liu2023grounding, ren2024grounded} to annotate open-world masks, but they still train the brush branch using entire image captions. CAT-Diffusion \cite{chen2025improving} uses a semantic inpainter to extract visual-text features and integrates them into the diffusion model, but the features extracted from the masked image and the short label text constrain the overall performance of the model. There is a strong need for a dataset containing mask and the detailed mask annotation pairs. To fill this gap, we propose MTAPipeline (Mask-Text Alignment) and MTADataset. For MTAPipeline, the input image is first annotated with the segmentation and recognition model to obtain object masks and labels. The masks and labels are then fed into the visual language model (VLM) to describe the labels and their styles in detail, resulting in detailed annotations of the masked objects. The MTADataset is built on MTAPipeline, consisting of 5 million images. Each image contains approximately 5 masks, and each mask has detailed descriptions of content and style. 

Structural distortion typically arises in generative models. Zhang et al. \cite{zhang2023adding} utilize ControlNet, which is a copy of the UNet encoder, to control the output using conditions such as pose or sketch. Specifically, CNI \cite{zhang2023adding} uses a mask as a condition for the inpainting task. With the consideration that inpainting task requires pixel-to-pixel constraints instead of sparse structural control, Xuan et al. \cite{ju2024brushnet} propose BrushNet for image generation using masks and masked images. Nevertheless,  it still suffers from structural distortion as shown in Figure ~\ref{fig:issues}. We argue that this is because the mask condition is not strong enough to control the inner structure of the object. To enhance the generated structure stability of the inpainting model, we propose a novel multi-task training strategy that jointly trains the inpainting task and edge prediction task. In this manner, the edge prediction task serves as a complementary mechanism that guides the inpainting model to reconstruct the content with stable structural features.

Even if the concept and structure of the generated object is reasonable, it is essential that the generated local objects are consistent with the original image in terms of style, including color tone, lighting conditions, and texture details. Gowthami et al. \cite{somepalli2024measuring} propose measuring style similarity using the output vectors of ViT \cite{alexey2020image}. StyleDiffusion \cite{yang2023zero} and GOYA \cite{wu2023not} calculate content loss and style loss with CLIP \cite{radford2021learning} features. PHDiffusion \cite{lu2023painterly} uses a pre-trained VGG \cite{simonyan2014very} network to obtain the foreground feature for contrastive style loss. Motivated by these works, we propose the integration of self-attention mechanisms with an inpainting style-consistency loss. The self-attention mechanism is useful for extracting dense image features. The loss function is designed to measure the style discrepancy between the generated image and the original image. Technically, we use a pre-trained VGG network to extract style features and then calculate the style loss in the latent space. By minimizing this inpainting style-consistency loss during the training process, the model is encouraged to generate regions that are not only visually coherent but also stylistically consistent with the surrounding content of the original image.

We evaluated our method on BrushBench \cite{ju2024brushnet} and EditBench \cite{xie2023smartbrush}, and the results demonstrate that it achieves state-of-the-art performance in terms of semantic alignment, structural stability, and style consistency compared to other methods. The main contributions of MTADiffusion are as follows:

\begin{itemize} \item We propose the MTAPipeline for collecting mask-text alignment data. Based on the MTAPipeline, we introduce a new MTADataset comprising 5 million images and 25 million masks, each with detailed annotation descriptions. \item A multi-task training strategy is proposed to enhance the generated structure stability of inpainting model by jointly training the inpainting task and edge prediction task. \item We design the combination of self-attention mechanisms with an inpainting style-consistency loss to ensure that the inpainting model generates object that are both content coherent and style-consistent. \end{itemize}

\begin{figure*}
    \centering
    \includegraphics[width=0.8\textwidth]{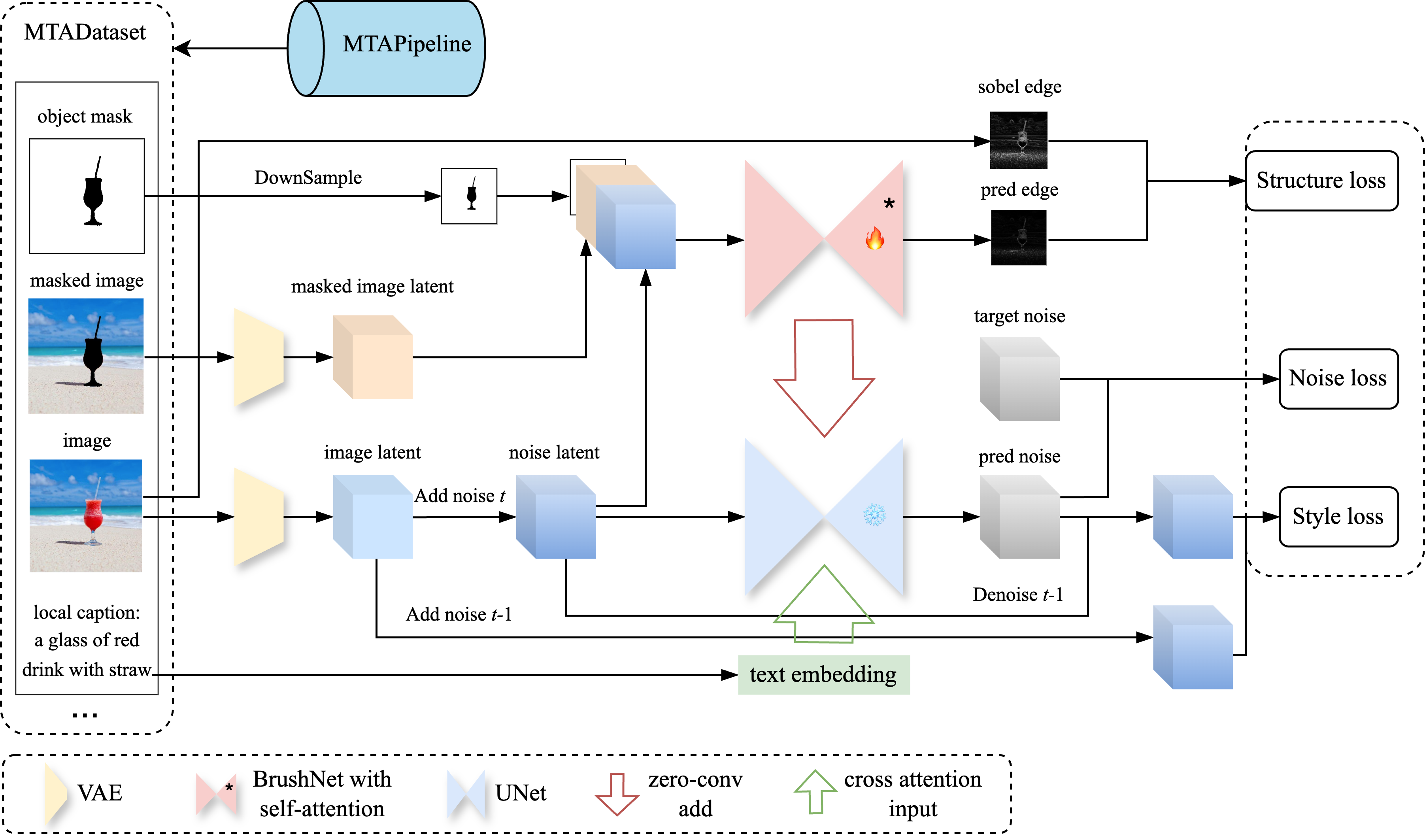}
    \caption{\textbf{The framework of MTADiffusion.} Our approach begins with constructing the MTAPipeline, which generates training samples for the MTADataset in the form of mask and text description pairs. The architecture of our network adopts a dual-branch strategy by applying the outputs of the upper branch to each layer of the UNet branch. To enhance the structural stability of the model, we jointly train the inpainting tasks and edge prediction tasks using a combination of noise loss and edge loss. The self-attention mechanisms and an inpainting style-consistency loss are designed to promote style consistency.}
    \label{fig:pipeline}
\end{figure*}

%% file: sec/2_related_work.tex
\section{Related Work}
\label{sec:related}

\subsection{Diffusion Models}
Diffusion models \cite{ho2020denoising, sohl2015deep} have attracted substantial attention for their superior performance in image synthesis. These models function by progressively adding noise into an image and subsequently training a model to reconstruct the original signal through an inverse denoising process. Stable Diffusion \cite{rombach2022high} marked a significant advancement, which involves training Diffusion Denoising Probabilistic Models \cite{ho2020denoising} in a lower-dimensional latent space learned via Variational Auto-Encoders \cite{kingma2013auto}. This approach not only accelerates the training and inference processes, but also preserves high image fidelity. Recent advances, particularly the SD-XL \cite{podell2023sdxl} model that leverages dual encoder architectures and an enlarged UNet architecture, have optimized the generative capacity. Moreover, the enhancements in SD3 \cite{esser2024scaling} that incorporate the T5 \cite{raffel2020exploring} model into the text encoder and substitute the UNet with a DiT \cite{peebles2023scalable} model significantly increase the output quality. These approaches have found widespread application across diverse domains, including unconditional image generation \cite{ho2020denoising, song2020denoising, song2019generative}, text-to-image synthesis \cite{rombach2022high, ramesh2022hierarchical, ruiz2023dreambooth}, and image editing \cite{couairon2022diffedit, hertz2022prompt}.

\subsection{Text-Guided Image Inpainting}

For image inpainting, the model incorporates an additional mask input to specify the modified region. RePaint \cite{lugmayr2022repaint} is a training-free method that iteratively introduces noise and denoises the masked region while keeping the unmasked region unchanged. BLD \cite{avrahami2023blended} blends masked and unmasked features in the latent space and employs progressive mask shrinking for larger masks. These methods often suffer from inconsistencies between the generated content and the background context.

SDI \cite{rombach2022high} enhances the inpainting capabilities by fine-tuning the UNet of Stable Diffusion \cite{rombach2022high} with incorporating the image, masked image, and mask as inputs. CNI \cite{zhang2023adding} trains ControlNet using the mask as a conditioning input for inpainting tasks. BrushNet \cite{ju2024brushnet} further leverages the encoder and decoder of UNet in its brush branch to capture pixel-to-pixel information. SmartBrush \cite{xie2023smartbrush} guides object inpainting using text and shape, training text-to-image and inpainting tasks simultaneously by treating the whole mask as a special case. PowerPaint \cite{zhuang2023task} categorizes inpainting tasks into 4 different subtasks and trains distinct prompts for each subtask. CATDiffusion \cite{chen2025improving} cascades a Transformer-based semantic inpainter and an object inpainting diffusion model to improve the semantic ability for text-guided object inpainting. Imagen Editor \cite{wang2023imagen}, based on Imagen \cite{saharia2022photorealistic}, optimizes inpainting through multi-level feature fusion. Additionally, there are multimodal inpainting models such as UniPaint \cite{yang2023uni}, a method that uses text, stroke, and exemplar as conditions to guide the inpainting process. Despite these advancements, these methods do not strictly align masks and textual descriptions during training, which limits the models' semantic understanding capabilities.

\subsection{Style Transfer}
In style transfer tasks, the training loss is generally disentangled into content loss and style loss. Style loss involves extracting style features such as CLIP \cite{radford2021learning} features, ViT \cite{alexey2020image} embeddings, or VGG \cite{simonyan2014very} features. Yang et al.\cite{yang2023zero} proposed a zero-shot contrastive loss for image style transfer, where their style loss uses CLIP loss to calculate the cosine distance in the CLIP embedding space between the generated image and the style prompt. StyleDiffusion\cite{wang2023stylediffusion} fine-tunes the style transfer module via a CLIP-based style disentanglement loss coordinated with a style reconstruction prior to implicitly learning disentangled style information. Gowthami et al.\cite{somepalli2024measuring} present a framework for understanding and extracting style descriptors from images by calculating the cosine similarity of two vectors obtained from ViT outputs. InST\cite{zhang2023inversion} learns the artistic style directly from a single painting and guides synthesis without requiring complex textual descriptions. GOYA\cite{wu2023not} uses CLIP and linear layer to calculate content loss and style loss. PHDiffusion\cite{lu2023painterly} employs content loss, AdaIN \cite{huang2017arbitrary} style loss, and contrastive style loss to balance the trade-off between style migration and content preservation. Inspired by these works, we use the VGG network to extract style features and then compute style loss with the Gram matrix.

%% file: sec/3_method.tex
\section{Method}

In this work, we propose MTADiffusion, an inpainting model with the superiority of semantic alignment, structural stability, and style consistency. The overall framework of our model is illustrated in Figure \ref{fig:pipeline}. Our approach begins with constructing the MTAPipeline, which generates training samples for the MTADataset in the form of mask and text description pairs. Similarly to BrushNet \cite{ju2024brushnet}, we adopt a dual-branch strategy by applying the outputs of the brush branch to each layer of the UNet branch. Inpainting tasks and edge prediction tasks are jointly trained using the combination of noise loss and edge loss to enhance the structural stability of the inpainting model. To mitigate style inconsistencies, we introduce self-attention mechanisms in the brush branches, and an inpainting style-consistency loss is calculated in the latent space.

In the following sections, we will present the key components of our approach step by step. First, we introduce the MTAPipeline and MTADataset in Section \ref{31}. Next, we describe our multi-task training strategy in Section \ref{32}. Following that, we explain the self-attention mechanisms and the inpainting style-consistency loss in Section \ref{33}. Finally, Section \ref{34} outlines the end-to-end training method employed to optimize our model.

\subsection{MTAPipeline and MTADataset}
\label{31}

\begin{figure}
    \centering
    \includegraphics[width=0.45\textwidth]{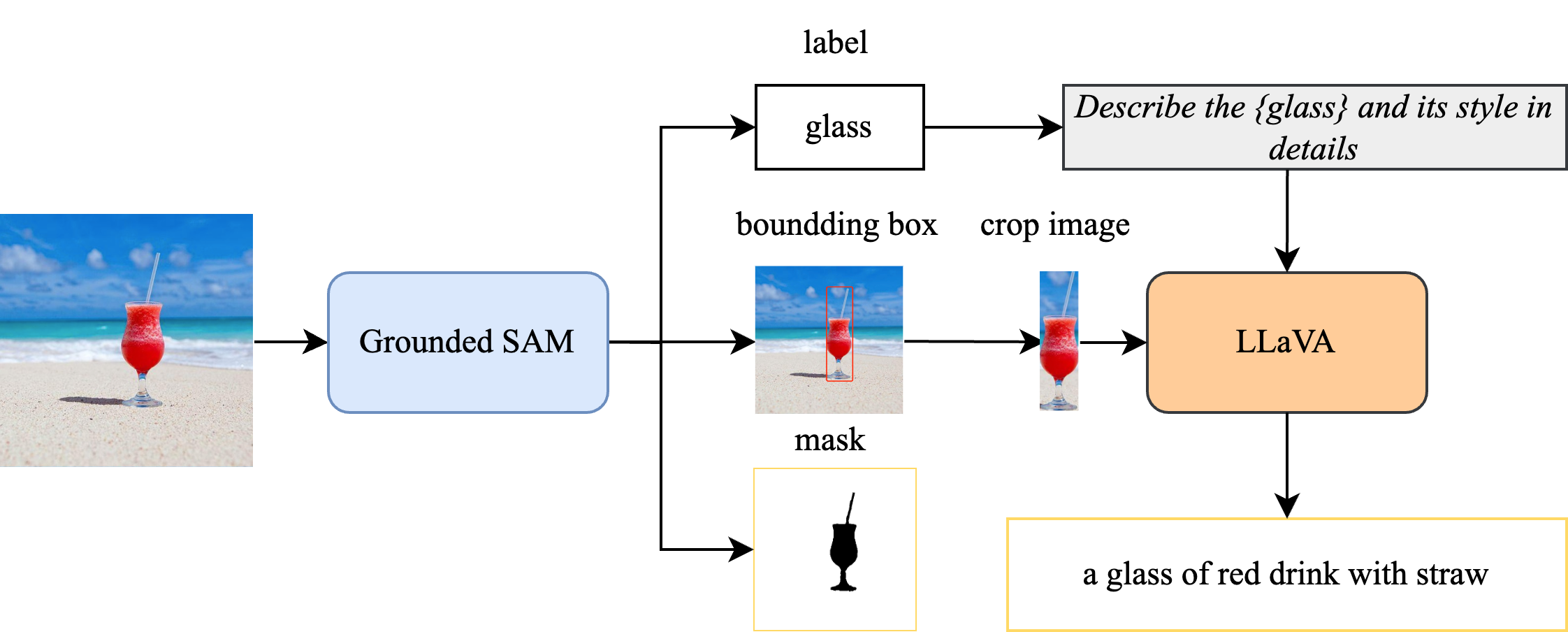}
        \caption{\textbf{MTAPipeline for annotating MTADataset.} We first employ Grounded-SAM \cite{ren2024grounded} to extract labels, bounding boxes, and masks. Subsequently, we utilize LLaVA \cite{liu2024visual} to annotate mask region with detailed content and style information.}
    \label{fig:llava_pipe}
\end{figure}

Our dataset is constructed from a subset of the LAION \cite{schuhmann2022laion} collection, comprising 5 million images. We selected images with an aesthetic score greater than 5.8 and the resolution larger than 1024 to obtain high-quality image sets. In the LAION dataset, each image is paired with a prompt that encapsulates its global information. However, for the inpainting task, precise content and style descriptions of local mask regions are essential, as user prompts in most cases specify their desired object without extensively describing the background information. 

Thus, we present the MTAPipeline, as illustrated in Figure \ref{fig:llava_pipe}. In the first stage, we obtain masks, noun labels, and bounding boxes using Grounded-SAM \cite{ren2024grounded}. Masks with a recognition confidence greater than 0.6 are selected as the final results. This process takes approximately 0.14 seconds per image and required around 6 days when executed on 8 V100 GPUs. In the second stage, we employ LLaVA \cite{liu2024visual}, which is an advanced vision language model, as the caption model to annotate local mask objects with text descriptions. The prompt \textit{"Describe the \{label\} and its style in details"} is used to describe the object and its corresponding style while ensuring the text stays within a manageable length. This caption stage was conducted over 2.5 days on 500 NVIDIA inference GPUs. The effects of different caption models will be discussed in the experiments detailed in Section \ref{52}.

Based on MTAPipeline, we constructed the MTADataset, which consists of 5 million images. Each image contains approximately 5 masks, and each mask is accompanied by detailed descriptions of its content and style.

\subsection{Joint Training for Inpainting and Edge Prediction Tasks}
\label{32}
For the structural distortion problem, we argue that this arises from the random noise not being effectively guided by the current denoising mechanism. To mitigate this, we introduce edge prediction tasks into the denoising process.
Technically, as shown in Figure ~\ref{fig:pipeline}, for the attention block in the brush branch, we extend the output dimension of the last layer by one additional dimension to predict the edge map. This changes the output dimension from \( (b, c, h, w) \) to \( (b, c+1, h, w) \), where \( b \) is the batch size, \( c \) is the number of channels, and \( h \) and \( w \) are the resolutions. Consequently, we obtain the predicted edge map \( s_{\text{pred}} \) with dimensions \( (b, 1, h, w) \).

For the input image \( I \), the Sobel edge extraction function \( f_{\text{sobel}} \) can be applied to compute the ground-truth edge map \( f_{\text{sobel}}(I) \). After applying a downsampling operation, we obtain the following:

\begin{equation}
\tilde{s} = \text{downsample}(f_{\text{sobel}}(I))
\end{equation}

Subsequently, the edge prediction loss can be defined as:

\begin{equation}
L_{\text{structure}} = \frac{1}{B} \sum_{i=1}^{B} \left\| s_{\text{pred}}^i - \tilde{s}^i \right\|_F^2
\label{structure}
\end{equation}

where \( B \) is the number of samples, and \( s_{\text{pred}}^i \) and \( \tilde{s}^i \) represent the predicted edge map and the downsampled ground-truth edge map for the \( i \)-th sample, respectively.

\subsection{Style-consistency Improvement}
\label{33}
We employ two strategies to enhance the style consistency between the generated objects and the background images. First, a self-attention block capable of incorporating mask information is integrated layer by layer with the outputs of the UNet. Second, a VGG-based latent style extractor is used to capture style features from different noise levels in the UNet outputs. This extracted style features guides the model in generating objects with greater style consistency through an end-to-end Gram loss.

\subsubsection{Self-attention for brush branch}
Similar to BrushNet \cite{ju2024brushnet}, we utilize a dual-branch strategy in the network, comprising a UNet branch and a brush branch. In contrast to  BrushNet, which replicates the residual blocks from the UNet architecture, our approach utilizes multi-resolution self-attention blocks. This design is motivated by the aim of enabling the brush branch network to attend to more global image information during the reconstruction process, thereby enhancing its ability to perceive the overall style of the image.

The weights of the brush branch are cloned from the pre-trained UNet model, excluding its cross-attention layers. The brush branch and the UNet branch are connected using zero convolutions, as shown in Equation ~\ref{attn}. 

\begin{equation}
\begin{aligned}
\epsilon_{\theta}\left(z_t, t, C\right)_j &= \epsilon_{\theta}\left(z_t, t, C\right)_j \\
&+ w \cdot \mathcal{Z}\left(\epsilon_{\theta}^{attn_j}\left(\left[z_t, z_0^{masked}, m^{resized}\right], t\right)_j\right)
\end{aligned}
\label{attn}
\end{equation}

where \( \epsilon_{\theta}\left(z_t, t, C\right)_j \) is the feature of the \textit{j}-th layer in UNet branch \( \epsilon_{\theta} \), \( C \) is the condition input.  \( \epsilon_{\theta}^{attn_j} \) is the feature in the brush branch with self-attention, which takes the noisy latent \( z_t \), masked image latent \( z_0^{masked} \) and downsampled mask \( m^{resized} \) as input. \( \mathcal{Z} \) is the zero convolution and \( w \) is the hyperparameter.

Then, the loss of ground truth noise \(\tilde{z}_i\) and pred noise \(z_i\) can be computed as follows:

\begin{equation}
L_{\text{noise}} = \frac{1}{B} \sum_{i=1}^{B} \left\| z_i - \tilde{z}_i \right\|_F^2
\label{noise}
\end{equation}

\subsubsection{Inpainting style loss}

To further enhance the model's capacity for style consistency, we utilize the VGG \cite{simonyan2014very} network to extract style features from the UNet outputs in the latent space. For the noise \( z_t \) predicted by the UNet, the denoised latent map at the \( t-1 \) time step of noise scheduler can be obtained through the denoise function \( f_{\text{denoise}}\):
\begin{equation}
    X_{t-1} = f_{\text{denoise}}(X_t, z_t, t)
\end{equation}

where \( z_t \) is the predicted noise and \( X_t \) is the noise latent at time \( t \).

Similarly, for the latent representation of the input ground truth image after encoding with VAE \cite{kingma2013auto}, we can also obtain the ground truth noise at time step \( t-1 \) by applying the noise addition function \(f_{\text{add\_noise}} \). 
\begin{equation}
    \tilde{X}_{t-1} = f_{\text{add\_noise}}(X_0, z_0, t-1)
\end{equation}

where \( X_0 \) is the latent representation and \( z_0 \) is the initial noise.
We then pass \( X_{t-1}, \tilde{X}_{t-1} \)through the VGG network to obtain the corresponding style embeddings \( (\alpha_1, \ldots, \alpha_n) \) and \( (\beta_1, \ldots, \beta_n) \)

Then our inpainting style loss can be calculated using the Gram matrices and mean squared error (MSE):

\begin{equation}
    L_{\text{style}} = \frac{1}{BN} \sum_{i=1}^{B} \sum_{i=1}^{N} \left\| G(\alpha_i) - G(\beta_i) \right\|_F^2
    \label{style}
\end{equation}

where \( G(\cdot) \) denotes the Gram matrix computation and \( N \) is the number of style features.

\begin{table*}[!ht]
    \centering
    \caption{\textbf{Quantitative comparisons to other diffusion-based inpainting models in BrushBench.}}
    \begin{tabular}{c|cc|ccc|cc}
        \hline
        \textbf{Metrics} & \multicolumn{2}{c|}{\textbf{Image Quality}} & \multicolumn{3}{c|}{\textbf{Masked Region Preservation}} & \multicolumn{2}{c}{\textbf{Text Alignment}} \\ \hline
        \textbf{Models} & \textbf{IR} $\times 10 \uparrow$ & \textbf{AS} $\uparrow$ & \textbf{PSNR} $\uparrow$ & \textbf{LPIPS} $\times 10^3 \downarrow$ & \textbf{MSE} $\times 10^3 \downarrow$ & \textbf{CLIP Sim} $\uparrow$ & \textbf{VQA Score} $\times 10^2 \uparrow$ \\ \hline
        SDI \cite{rombach2022high} & 11.72 & 6.50 & 21.52  & 48.39 & 13.87 & 26.17 & 64.55  \\ \hline
        PP \cite{zhuang2023task} & 11.46 & 6.24 & 21.43 & 48.43 & 32.73 & 26.48  & 66.50 \\ \hline
        CNI \cite{zhang2023adding} & 11.21 & 6.39 & 22.73 & 43.49 & 24.58 & 26.22  & 66.06 \\ \hline
         BrushNet \cite{ju2024brushnet} & 12.52 & 6.37 & 31.82 & 18.95 & 0.82 & 26.32 & 68.22 \\ \hline
        \textbf{Ours} & \textbf{12.69} & \textbf{6.50} & \textbf{31.87} & \textbf{18.94} & \textbf{0.80} & \textbf{26.52} & \textbf{68.97} \\ \hline
    \end{tabular}
    \label{tab:comparisonb}
\end{table*}

\begin{table*}[!ht]
    \centering
    \caption{\textbf{Quantitative comparisons to other diffusion-based inpainting models in EditBench.}}
    \begin{tabular}{c|cc|ccc|cc}
        \hline
        \textbf{Metrics} & \multicolumn{2}{c|}{\textbf{Image Quality}} & \multicolumn{3}{c|}{\textbf{Masked Region Preservation}} & \multicolumn{2}{c}{\textbf{Text Alignment}} \\ \hline
        \textbf{Models} & \textbf{IR} $\times 10 \uparrow$ & \textbf{AS} $\uparrow$ & \textbf{PSNR} $\uparrow$ & \textbf{LPIPS} $\times 10^3 \downarrow$ & \textbf{MSE} $\times 10^3 \downarrow$ & \textbf{CLIP Sim} $\uparrow$ \\ \hline
        SDI \cite{rombach2022high} & 1.86 & 5.69 & 23.25 & 24.30 & 6.94 & 28.00 \\ \hline
        PP \cite{zhuang2023task} & 1.24 & 5.44 & 23.34 & 24.12 & 20.12 & 27.80 \\ \hline
        CNI \cite{zhang2023adding} & 0.90 & 5.46 & 22.61 & 26.14 & 35.93 & 27.74  \\ \hline
         BrushNet \cite{ju2024brushnet} & 4.46 & 5.82 & \textbf{33.66} & 10.12 & \textbf{0.63} & 28.87 \\ \hline
        \textbf{Ours} & \textbf{4.82} & \textbf{5.85} & 33.31 & \textbf{10.10} & 0.65 & \textbf{29.12} \\ \hline
    \end{tabular}
    \label{tab:comparisone}
\end{table*}

\subsection{End-to-End Training for Style, Content, and Structural Consistency}
\label{34}
During training, we integrate these three loss functions to perform end-to-end training. Combining Equations ~\ref{structure}, \ref{noise} and \ref{style}, the final end-to-end loss can be computed as:
\begin{equation}
    L = \gamma L_{\text{noise}} + \delta L_{\text{style}} + \eta L_{\text{structure}} 
\end{equation}
where \( \gamma \), \( \delta \), and \( \eta \) are the weighting hyper-parameters.

The weights of the VAE and UNet are kept frozen, allowing the model to focus on the training of the newly introduced attention branch and the VGG latent style extractor simultaneously. Consequently, the model can effectively balance the integration of content coherence and style consistency, leading to improved outcomes in inpainting tasks.

%% file: sec/4_exp.tex
\section{Experiments}

\begin{table*}
    \centering
    \caption{\textbf{Comparison of Different Datasets.} We trained BrushNet for 10,000 iterations on both BrushData and our dataset, and observed significant improvements in image quality and text consistency in our dataset.}
    \begin{tabular}{c|cc|ccc|c}
        \hline
        \textbf{Metrics} & \multicolumn{2}{c|}{\textbf{Image Quality}} & \multicolumn{3}{c|}{\textbf{Masked Region Preservation}} & \multicolumn{1}{c}{\textbf{Text Alignment}} \\ \hline
        \textbf{Datasets} & \textbf{IR} $\times 10 \uparrow$ & \textbf{AS} $\uparrow$ & \textbf{PSNR} $\uparrow$ & \textbf{LPIPS} $\times 10^3 \downarrow$ & \textbf{MSE} $\times 10^3 \downarrow$ & \textbf{CLIP Sim} $\uparrow$ \\ \hline
        BrushData \cite{ju2024brushnet} & 11.73 & 6.33 & 30.90 & 20.06 & 0.98 & 26.28  \\ \hline
        MTADataset & 12.08 & 6.37 & 30.74 & 19.03 & 0.82 & 26.41  \\ \hline
    \end{tabular}
    \label{tab:diff_datasets}
\end{table*}

\begin{table*}
    \centering
    \caption{\textbf{Comparison of Different Caption Models.} The results after training for 10,000 iterations using different types of annotations as textual prompts.}
    \begin{tabular}{c|cc|ccc|cc}
        \hline
        \textbf{Metrics} & \multicolumn{2}{c|}{\textbf{Image Quality}} & \multicolumn{3}{c|}{\textbf{Masked Region Preservation}} & \multicolumn{1}{c}{\textbf{Text Alignment}} \\ \hline
        \textbf{Prompt Type} & \textbf{IR} $\times 10 \uparrow$ & \textbf{AS} $\uparrow$ & \textbf{PSNR} $\uparrow$ & \textbf{LPIPS} $\times 10^3 \downarrow$ & \textbf{MSE} $\times 10^3 \downarrow$ & \textbf{CLIP Sim} $\uparrow$  \\ \hline
        Original Image Caption & 11.67 & 6.37 & 30.21 & 20.92 & 1.13 & 26.31 \\ \hline
        BLIP2 Caption \cite{li2023blip}   & 11.49 & 6.33 & 30.62 & 20.41 & 1.02 & 26.28 \\ \hline
        Grounded-SAM Label \cite{ren2024grounded} & 11.41 & 6.34 & 30.66 & 20.40 & 1.02 & 26.23 \\ \hline
        LLaVA Caption \cite{liu2024visual}  & 11.76 & 6.36& 30.66 & 20.38 & 1.01 & 26.36 \\ \hline
    \end{tabular}
    \label{tab:diff_prompt}
\end{table*}

\subsection{Implementation Details}
\label{41}
In the experiments, the hyper-parameters \( \gamma \), \( \delta \), and \( \eta \) were set to 1, 100, and 0.1, respectively. The learning rate for the attention branch was set to \( 1 \times 10^{-5} \), while the learning rate for the VGG latent style extractor is \( 1 \times 10^{-7} \). The model was trained utilizing 8 NVIDIA V100 GPUs, achieving a total of 200,000 iterations.

For the VGG feature extractor, we increased the number of channels in the input layer from 3 to 4 to match the latent channels. The outputs of \( conv_{i\_1} \), where the \( i = (1,\ldots,5 )\) layers were considered as style features. During training, we use the caption of the entire image as the prompt for random masks and the detailed description of the object as the prompt for object-specific masks. To this end, we unify the training process within a single model to handle both random masks and object masks concurrently.

\subsection{Datasets}
\label{42}
Our training dataset, MTADataset, as described in Section~\ref{31}, is a self-annotated collection of 5 million images. For each image, the MTAPipeline generates approximately 5 masks, with each mask being annotated with a text label.

We evaluate our model on BrushBench \cite{ju2024brushnet} and EditBench \cite{wang2023imagen}. BrushBench consists of 600 images with object masks and image captions. The dataset includes a distribution of natural and artificial images, and is balanced across categories like humans, animals, indoor scenes, and outdoor scenes. EditBench consists of a collection of 240 images with random masks and captions, with an equal ratio of natural images and generated images.

\subsection{Evaluation Metrics}
\label{43}
We evaluate the performance of our model from three aspects:

\begin{itemize} 
\item \textbf{Image Quality}: We assess the aesthetic appeal of the generated images using Image Reward(IR) \cite{xu2024imagereward} and Aesthetic Score(AS) \cite{schuhmann2022laion}. These metrics can also reflect the style and structural consistency of the images to some extent. Intuitively, images with greater consistency in style and structure are likely to exhibit higher aesthetic appeal.

\item \textbf{Masked Region Reconstruction}: We evaluate the reconstruction quality of the masked areas using LPIPS \cite{zhang2018unreasonable}, PSNR, and MSE metrics.

\item \textbf{Semantic Consistency}: To assess the semantic ability of the inpainting model, we employ the global CLIP Score \cite{wu2021godiva} to calculate the similarity between the entire image and the prompt. Furthermore, we use the VQA score \cite{lin2024evaluating} to accurately measure the similarity between the masked region and the prompt.
\end{itemize}

\subsection{Quantitative Results}
\label{44}
The evaluation results on BrushBench are presented in Table \ref{tab:comparisonb}. We compare our method with SDI \cite{rombach2022high}, CNI \cite{zhang2023adding}, PowerPaint \cite{zhuang2023task} and BrushNet \cite{ju2024brushnet}. Note that the BrushNet metrics were tested using the released model. As shown in the results, our method demonstrates superior performance in both semantic alignment and overall image quality. Specifically, our model outperforms other methods on the local VQA Score metrics, indicating that our dataset significantly enhances the model's semantic capabilities. Moreover, the IR and AS metrics reflect the structure and style superiority of our model. For the masked region preservation, we employed the same blending strategy as BrushNet, resulting in metrics that are close to those of it.

The performance in EditBench is shown in Table \ref{tab:comparisone}, which also demonstrates the superior performance of our method. It is noteworthy that our model addresses both object mask and random mask tasks simultaneously.

\subsection{Qualitative Results}
\label{45}

\begin{figure*}
    \centering
    \includegraphics[width=0.8\textwidth]{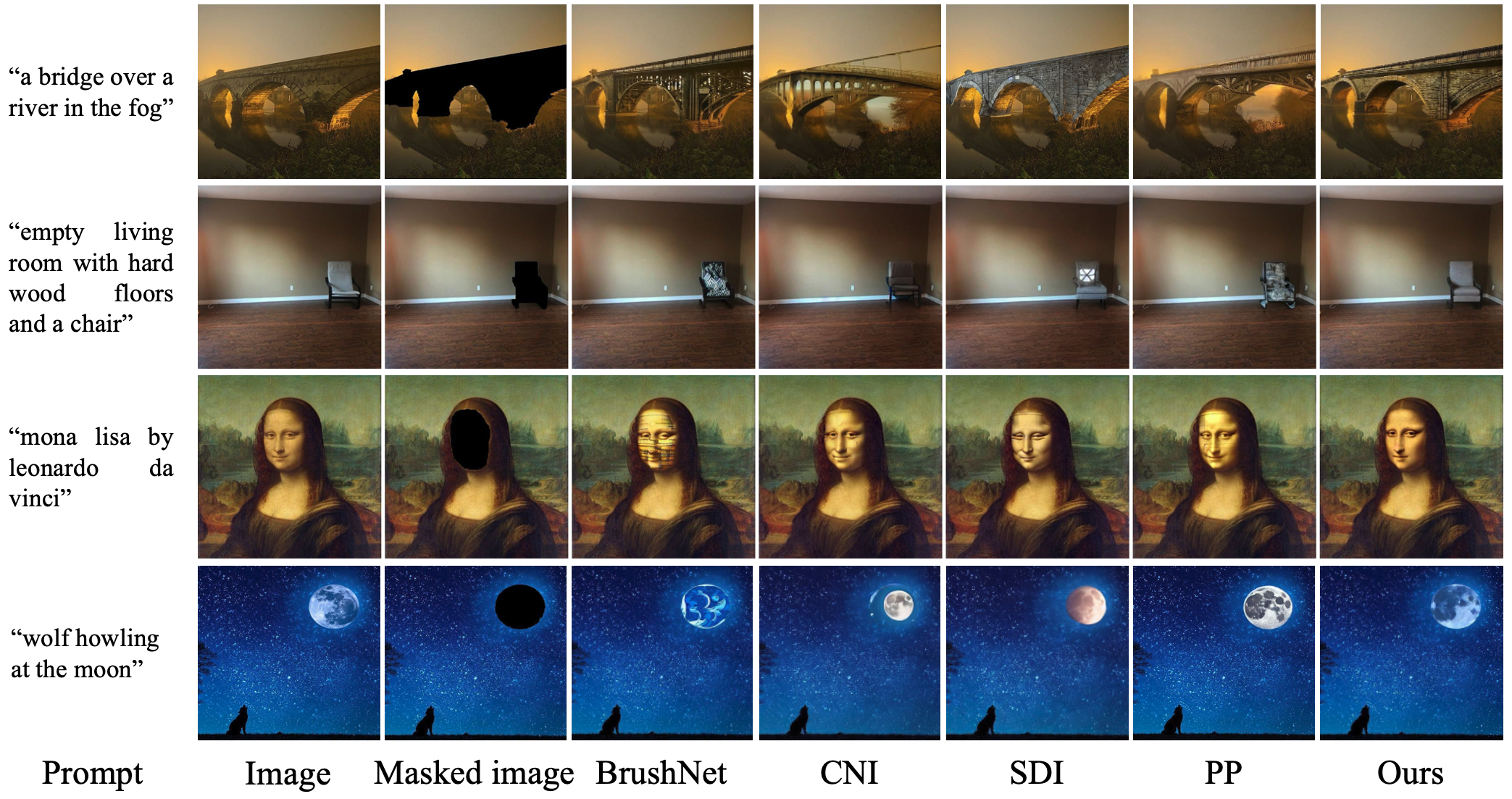}
    \caption{\textbf{Results on BrushBench compared to other methods.} With our method, the inpainting region achieves effective improvements in semantic alignment, structural stability, and style consistency.}
    \label{fig:results_b}
\end{figure*}

\begin{figure*}[!h]
    \centering
    \includegraphics[width=0.8\textwidth]{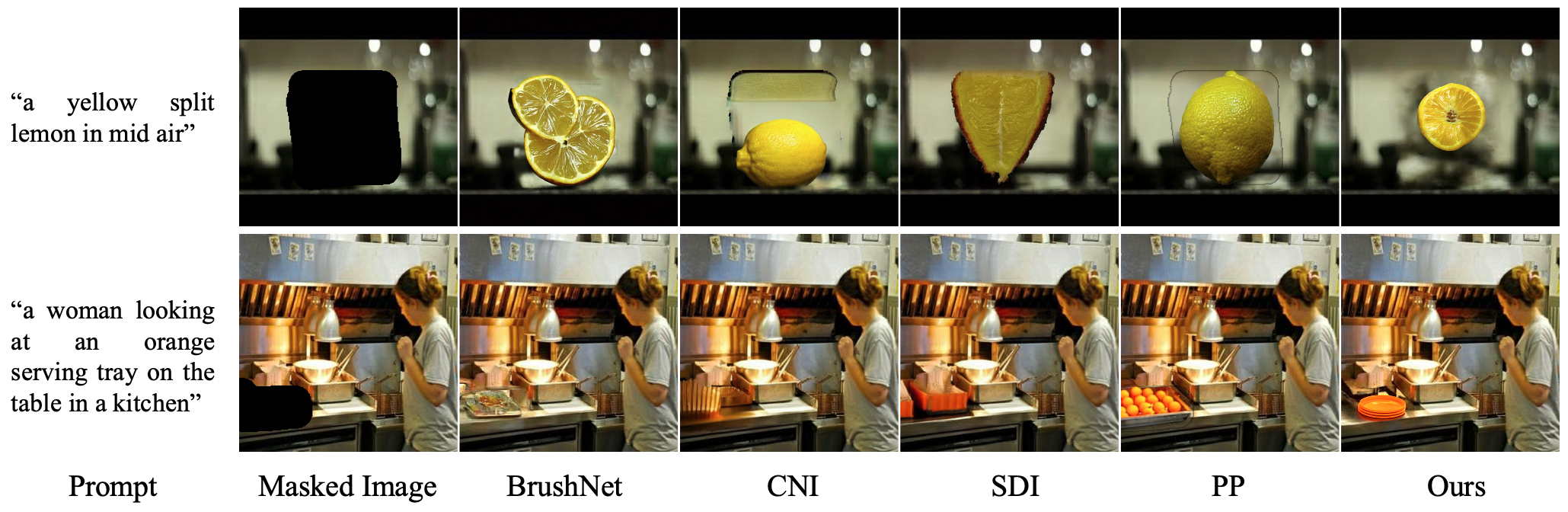}
    \caption{\textbf{Results on EditBench compared to other methods.} Our model trained on MTADataset demonstrates stronger semantic capabilities under random masking conditions.}
    \label{fig:results_e}
\end{figure*}

Figures \ref{fig:results_b} and Figure \ref{fig:results_e} present a qualitative comparison with other image inpainting methods on BrushBench and EditBench. We display results on natural images for random mask inpainting. As we can see, SDI and CNI exhibit poorer semantic ability as they were trained using image captions and random masks, causing semantic misalignment during the training stage. Our model, trained with mask-text pairs, shows significantly improved semantic performance in scenarios involving random masks. 

It is worth mentioning that SDI, which was trained on 600 million images, shows excellent style consistency with the generated content and the origin image. PowerPaint shows weaknesses in structural rationality. BrushNet exhibits style inconsistencies as its style is heavily influenced by the base model. The results of our proposed approach indicate that our method surpasses existing methods in semantic, structure and style. See also the supplementary material for more qualitative examples.

\subsection{User Study}
\label{46}

We randomly selected 30 images each from BrushBench and EditBench, resulting in a test set of 60 images that include both object masks and random masks. For each image, we presented the volunteers with the original image, the masked image, the prompt, and the inpainting results of 5 different methods, including SDI, CNI, PowerPaint, BrushNet, and Ours. To avoid biasing the user towards a specific approach, the results were shown in a different random order. The volunteers were then required to select one of the 5 images corresponding to 5 methods: \textit{Which image best matches the prompt? Which image has the most reasonable structure in the generated content? Which image shows the best style consistency between the generated content and the background?} These three questions are designed to evaluate the performance of different models in terms of semantic alignment, structural stability, and style consistency, respectively. A total of 30 participants took part in the study, providing 1800 valid votes for each question. We calculated the percentage of votes for each method. The statistical results are shown in Table \ref{tab:user_study}.
Obviously, our method performed the best.

\begin{table}[!h]
    \centering
    \caption{\textbf{Results of the user study.} We calculated the percentage of votes from volunteers for each method.}
    \resizebox{\linewidth}{!}{
    \begin{tabular}{c|c|c|c|c|c}
    \hline
        \textbf{Votes \%} & SDI & CNI  & PowerPaint & BrushNet  & Ours  \\ \hline
        Semantic alignment & 3 & 6 & 10 & 15 & \textbf{66} \\ \hline
        Structural stability & 3 & 7 & 14 & 16 & \textbf{60} \\ \hline
        Style consistency & 22 & 3 & 8 & 13 & \textbf{54} \\ \hline
    \end{tabular}}
    \label{tab:user_study}
    \vspace{-0.3cm} 
\end{table}

\section{Ablation Experiments}
In this section, we conduct a series of comparative experiments. First, we present dataset experiments to validate the impact of different training sets on the results, as detailed in Section \ref{51}. Next, we perform an ablation study on different caption models of MTAPipeline with the experimental outcomes discussed in Section \ref{52}. See also supplementary material for the ablation studies of different losses.

\subsection{Datasets}
\label{51}
The impact of different training datasets on model performance is illustrated in Table \ref{tab:diff_datasets}. We trained BrushNet for 10,000 iterations using both BrushData and MTADataset and evaluated the results on BrushBench. The results indicate significant improvements in image quality and text consistency when using our dataset. These enhancements can be attributed to the unique characteristics and annotations present in our dataset, which provide richer contextual information and a more diverse set of examples for the model to learn from. Consequently, the model demonstrates better alignment with prompt and improved visual fidelity in the generated images.

\subsection{Caption Models}
\label{52}

To evaluate the influence of different caption models used in the MTAPipeline, we employed various types of annotations as input text prompts during training, including original image captions from the LAION dataset \cite{schuhmann2022laion}, labels from Grounded-SAM \cite{ren2024grounded}, image captions generated by BLIP2 \cite{li2023blip}, and our annotated descriptions that incorporate style information using LLaVA \cite{liu2024visual}.

The results after training for 10,000 iterations in BrushBench are shown in Table \ref{tab:diff_prompt}. The original prompts yielded substantial image rewards and aesthetic scores, demonstrating their utility in generating reasonable output. In contrast, BLIP2-annotated prompts produced slightly lower image quality metrics, indicating a potential limitation in the descriptive accuracy of automatically generated captions. Simple labels derived from Grounded-SAM showed relatively poorer performance in terms of text alignment, suggesting that rich text description is essential for the semantic ability. Remarkably, our LLaVA annotations, which integrated the detailed text descriptions of the style and content information, achieved the best performance in all metrics.

%% file: sec/5_conclusion.tex
\section{Conclusion}

We present MTADiffusion to enhance the capabilities of the inpainting models in semantic alignment, structural stability, and style consistency using the MTAPipeline, a multi-task training strategy, and a novel style loss. Notably, these strategies are sufficiently generic to be applicable to other models for performance improvement. Meanwhile, we believe that the MTADataset has the potential to significantly advance research within the academic community. 

The limitation of our method is that the precision of the text descriptions depends on the caption model. In some cases, the descriptions of LLaVA are not completely accurate, such as the color of object, which limits the model’s ability of precise generation. Future work can further improve the quality of the dataset by leveraging more advanced segmentation and caption models. 

\section*{Acknowledgment}

Xiaolin Hu was supported by the National Natural Science Foundation of China (No. U2341228).